\documentclass[lettersize,journal]{IEEEtran}
\usepackage{amsmath,amsfonts}
\usepackage{algorithmic}
\usepackage{algorithm}
\usepackage{array}
\usepackage{subfigure}
\usepackage{textcomp}
\usepackage{booktabs}
\usepackage{stfloats}
\usepackage{url}
\usepackage{verbatim}
\usepackage{graphicx}
\usepackage{cite}
\usepackage{ulem}
\usepackage{balance}
\usepackage{color,xcolor}
\usepackage{amsthm}

\theoremstyle{definition}
\newtheorem{definition}{Definition}

\newtheorem{property}{Property}

\renewcommand{\emph}[1]{\textit{#1}}
\def\etal{\textit{et al.~}}
\def\S{\mathcal{S}}

\def\mnd{\mathrm{MND}}
\def\dtwh{\mathrm{DTW}\text{-}\mathrm{H}}
\newcommand{\ist}[1]{#1^\text{st}}

\newcommand{\ith}[1]{#1^\text{th}}
\newcommand{\towith}[1]{\mathop{\to}\limits^{#1}}

\begin{document}

\title{State Space Closure: Revisiting Endless Online Level Generation via Reinforcement Learning}

\author{
  Ziqi Wang,~\IEEEmembership{Student Member,~IEEE},
  Tianye Shu,
  Jialin Liu,~\IEEEmembership{Senior Member,~IEEE}
  \thanks{This work has been accepted by the IEEE Transactions on Games.}
  \thanks{
    The authors are with the Guangdong Provincial Key Laboratory of Brain-Inspired Intelligent Computation, Department of Computer Science and Engineering, Southern University of Science and Technology (SUSTech), Shenzhen 518055, China. They are also with the Research Institute of Trustworthy Autonomous System, SUSTech.}
      
% <-this % stops a space
 % \thanks{Manuscript received April 19, 2021; revised August 16, 2021.}
}

% The paper headers
% \markboth{Journal of \LaTeX\ Class Files,~Vol.~14, No.~8, August~2021}%
% {Shell \MakeLowercase{\textit{et al.}}: A Sample Article Using IEEEtran.cls for IEEE Journals}

%\IEEEpubid{0000--0000/00\$00.00~\copyright~2021 IEEE}
% Remember, if you use this you must call \IEEEpubidadjcol in the second
% column for its text to clear the IEEEpubid mark.

\maketitle

\begin{abstract}
  In this paper, we revisit endless online level generation with the recently proposed experience-driven procedural content generation via reinforcement learning (EDRL) framework.
  Inspired by an observation that EDRL tends to generate recurrent patterns, we formulate a notion of \textit{state space closure} which makes any stochastic state appeared possibly in an infinite-horizon online generation process can be found within a finite-horizon.
  Through theoretical analysis, we find that even though state space closure arises a concern about diversity, it generalises EDRL trained with a finite-horizon to the infinite-horizon scenario without deterioration of content quality. 
  Moreover, we verify the quality and the diversity of contents generated by EDRL via empirical studies, on the widely used \textit{Super Mario Bros.} benchmark.
  Experimental results reveal that the diversity of levels generated by EDRL is limited due to the state space closure, whereas their quality does not deteriorate in a horizon which is longer than the one specified in the training.
  Concluding our outcomes and analysis, future work on endless online level generation via reinforcement learning should address the issue of diversity while assuring the occurrence of state space closure and quality.
\end{abstract}


\section{Introduction}

\IEEEPARstart{A}{lthough} deep learning has achieved notable successes and shown its advantages, including efficiency, scalability and generalisation capability, in procedural content generation (PCG), endless online generation of game levels via deep learning remains to be a rarely explored direction~\cite{summerville2018procedural,liu2021deep}.
Recently, Shu \etal \cite{shu2021experience} introduce an experience-driven PCG via reinforcement learning (EDRL) framework, integrated generative adversarial networks (GAN) \cite{goodfellow2014gan, volz2018evolving} and deep reinforcement learning (RL) \cite{sutton2018reinforcement} to realise the online level generation of game levels while maximising experience-driven quality measurements. The work of \cite{shu2021experience} reveals the potential of endless online level generation (OLG) via deep learning. 
  
\begin{figure}[t]
    \centering
    \subfigure[Slices of a level generated by EDRL approach with $\gamma=0.70$, $n=4$.]{\includegraphics[width=\linewidth]{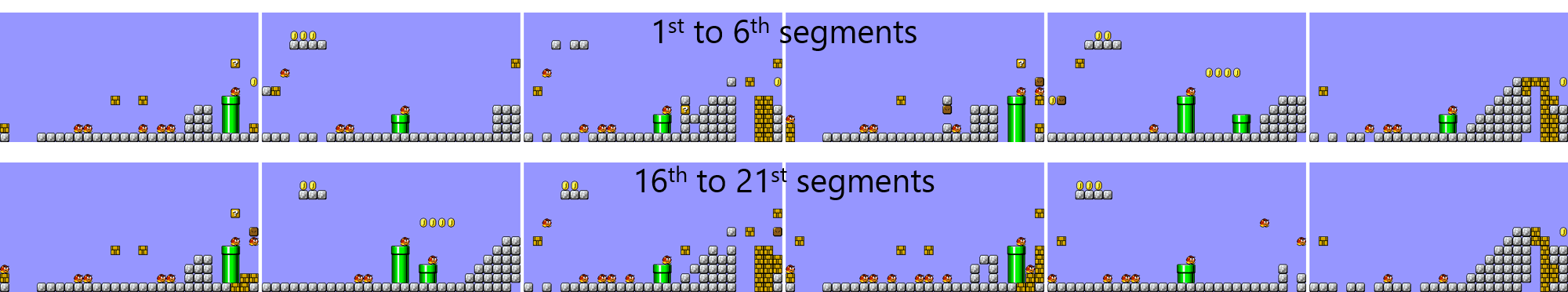}}
    \subfigure[Slices of a level generated by EDRL approach with $\gamma=0.99$, $n=6$.]{\includegraphics[width=\linewidth]{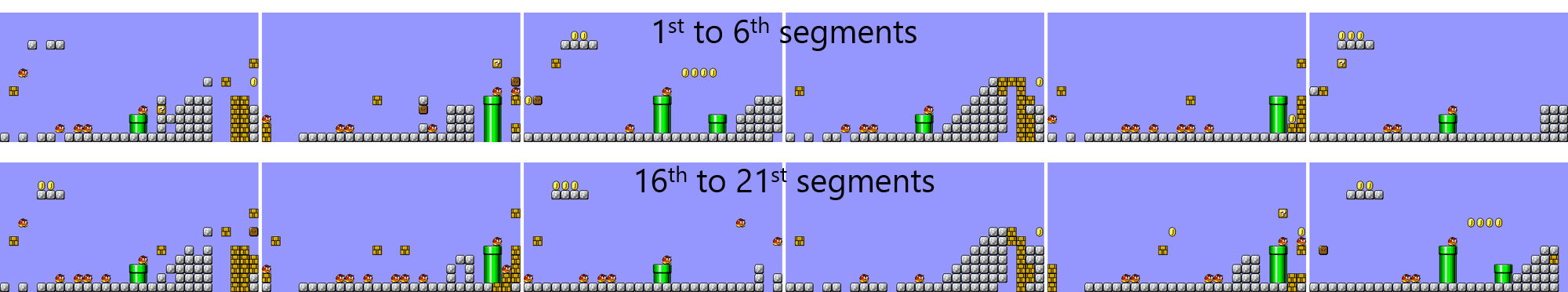}} 
    \caption{Two levels generated by EDRL approach, detailed in Section \ref{sec:exp}.}
    \label{fig:levels}
  \end{figure}

The work of \cite{wang2022fun} shows that EDRL can generate levels with promising short-term novelty (i.e., local intra-level divergence) with proper configuration of the reward function. However, we find that EDRL tends to generate recurrent patterns (cf. Fig. \ref{fig:levels}). This is undesired because if some patterns repetitively appear in levels, players will get bored easily. An in-depth understanding of this phenomenon will help improve the diversity of levels generated via EDRL.

To analyse this undesired yet interesting phenomenon, this work formally defines the \textit{state space closure} (SSC), which describes a scenario that the set of all possible states in an infinite-horizon Markov decision process (MDP) converges to a closure within a finite number of steps $h$. In other words, all states that possibly appear after $h$ steps of the OLG process can be found within the first $h$ steps. It can lead to poor long-term novelty and inter-level diversity, but it can also make RL designers always able to generate level segments as fine as during training, even if proceeding to infinite steps. 
  
In the rest of this paper, (i) we formally define SSC and introduce its mathematical property; (ii) the occurrence of SSC is verified with EDRL; (iii) we further measure the diversity of EDRL designer and verify that the content quality, in terms of rewards, will not decrease when generating a level longer than the one specified during training. Results show that the diversity of levels generated by the state-of-the-art EDRL is limited due to SSC\footnote{The code, trained models, results and plots of this work are available in \url{https://github.com/SUSTechGameAI/MFEDRL}.}. Future work should consider addressing the issue of diversity.

  \IEEEpubidadjcol

\section{Online Level Generation via RL}\label{sec:olgrl}
It is natural to model the OLG task as an MDP and apply RL to train online level generators \cite{khalifa2020pcgrl,shu2021experience}. However, previous works on procedural content generation via reinforcement learning (PCGRL)
\cite{khalifa2020pcgrl,werneck2020generating,susanto2021applying,zakaria2022procedural} typically model an action as a single-tile mutation, which makes it not efficient enough to realise real-time generation. Motivated by this, EDRL~\cite{shu2021experience} employs a trained GAN \cite{volz2018evolving} as an action decoder and models the action space of an RL level designer as the latent space of this GAN. At the $\ith{i}$ step of the MDP, the RL designer observes $n$ previous level segments in latent vector representation $s_i = z_{i-n} \oplus \cdots \oplus z_{i-1}$, where $\oplus$ denotes the vector concatenation, and then the RL designer decides an action $a_i$ (i.e., a latent vector $z_i$) as an input to the GAN (i.e., segment generator). The GAN decodes $a_i$ into a level segment to be appended to the current level.
The use of GAN enables efficient generation. Moreover, GAN learns to generate realistic level segments \cite{volz2018evolving} so that the basic content quality is assured in EDRL. In addition to the work of \cite{shu2021experience}, the work of \cite{wang2022online} proposes to generate game levels from music online via RL. Wang \etal extend EDRL to multi-faceted experienced-driven OLG~\cite{wang2022fun}. However, none of them  \cite{shu2021experience,wang2022online,wang2022fun} has discussed the recurrent patterns in generated levels. 
  
\section{State Space Closure}
  To understand the cause of recurrent patterns, this section defines the state space closure, analyses it theoretically, and then verifies its occurrence in the implementation of EDRL on the widely used \textit{Super Mario Bros.} (SMB) benchmark.
  \subsection{Definition of State Space Closure}
  In the context of OLG via RL described in Section \ref{sec:olgrl}, with $\S$ denotes the state space, i.e., all possible permutations of any $n$ latent vectors, $\pi$ denotes the RL policy (i.e., designer), $\S^\pi_i$ denotes the set of all possible states at the $\ith{i}$ step, $\S^\pi_{s:e} = \bigcup_{i=s}^e \S^\pi_i$ denotes the set of all possible states in the step interval $[s, e]$, we define the \emph{state space closure} as follows.
  \begin{definition}[State Space Closure, SSC]
  Consider an RL policy making decisions on an MDP with a horizon of $h$ (i.e., each episode terminates after $h$ steps), with an integer $g$ such that $g < h$, the SSC occurs at $[g, h]$ if $\S^\pi_h \subseteq \S^\pi_{g:h-1}$.
  \end{definition}

  Now consider the set of all possible states transited from $\S^\pi_i$ in one step with $\pi$, namely $\{ \S^\pi_i \towith{\pi} \cdot \}$. As an MDP satisfies the Markov property, i.e., the state $s_i$ only depends on the previous state $s_{i-1}$ and action $a_{i-1}$, we have $\S^\pi_i = \{ \S^\pi_{i-1} \towith{\pi} \cdot \}$. Similarly, we have $\S^\pi_{s:e} = \{ \S^\pi_{s-1:e-1} \towith{\pi} \cdot \}$. With a remark that if $A \subseteq B$ then $\{ A \towith{\pi} \cdot \} \subseteq \{ B \towith{\pi} \cdot \}$, where $A, B$ is any two subsets of $\S$, we formulate a useful property as below. 
  \begin{property}
    If SSC occurs at $[g, h]$, then the MDP will never transit to any state out of $\S^\pi_{g:h}$ even if the step number goes to infinite, i.e., $\S^\pi_{h:\infty} = \S^\pi_{g:h-1}$.
  \label{property}
  \end{property}
  \begin{proof}
  \begin{equation}
    \allowdisplaybreaks
    \begin{aligned}
      & ~\S^\pi_h \subseteq \S^\pi_{g:h-1} \\
      \Rightarrow & ~ \{ \S^\pi_{h} \towith{\pi} \cdot \} \subseteq \{ \S^\pi_{g:h-1} \towith{\pi} \cdot \} \Rightarrow \S^\pi_{h+1} \subseteq \S^\pi_{g+1:h} \\
      \Rightarrow & ~ \S^\pi_{h+1} \subseteq (\S^\pi_{g+1:h-1} \cup \S^\pi_h) = \S^\pi_{g+1:h-1} \subseteq \S^\pi_{g:h-1} \\
      \Rightarrow & ~ \S^\pi_{h+2} \subseteq \S^\pi_{g:h-1} \Rightarrow \cdots \\
      \Rightarrow & ~ \S^\pi_{h:\infty} \subseteq \S^\pi_{g:h-1}.
    \end{aligned}
    \label{eq:ssc}
  \end{equation}
  \end{proof}
  
  The derivation of Eq. \eqref{eq:ssc} shows that if no unseen state will be introduced at the $\ith{h}$ step, then the MDP will never transit to any unseen state even if the step number goes to infinite. This can imply a poor inter-level diversity, which depends on the scale of $\S^\pi_{g:h-1}$.
  It also indicates that the RL policy can learn to make decisions at any state that possibly appears in an endless generation, even if it was trained in an infinite horizon of $h$. Therefore, we can train a robust endless online level generator with a finite-horizon MDP if SSC occurs. 

  \subsection{Occurrence of State Space Closure}
\begin{figure}[t]
    \centering
    \includegraphics[width=0.45\linewidth]{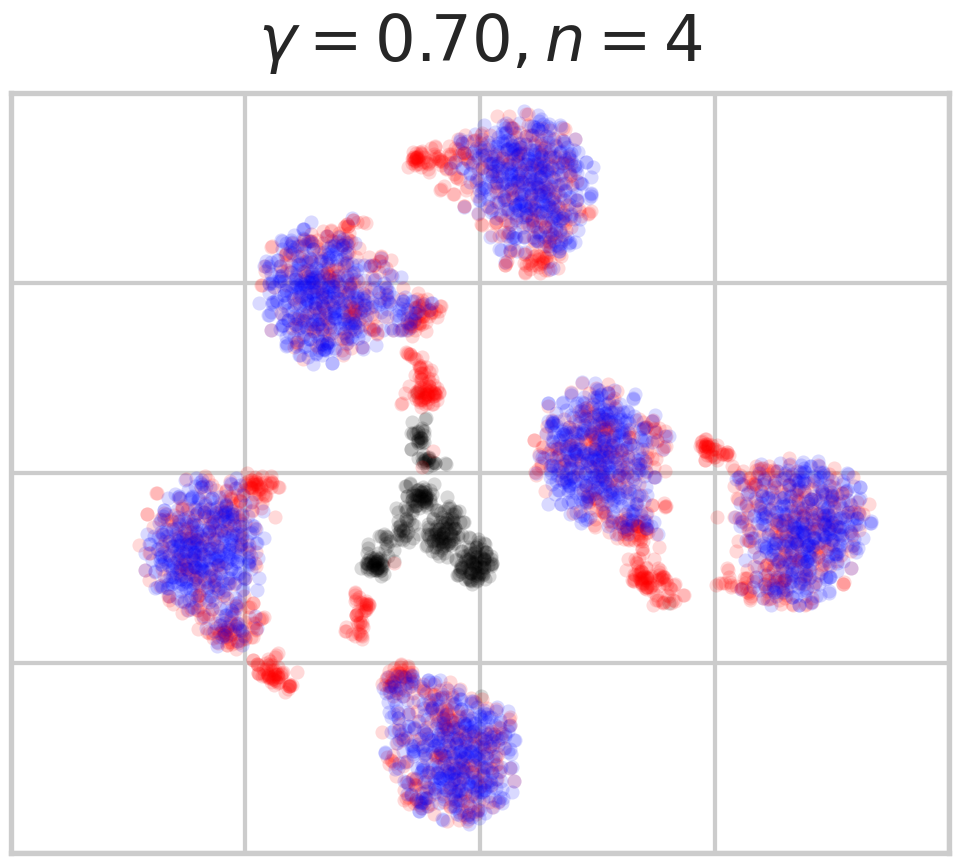} 
    \hspace{6pt}
    \includegraphics[width=0.45\linewidth]{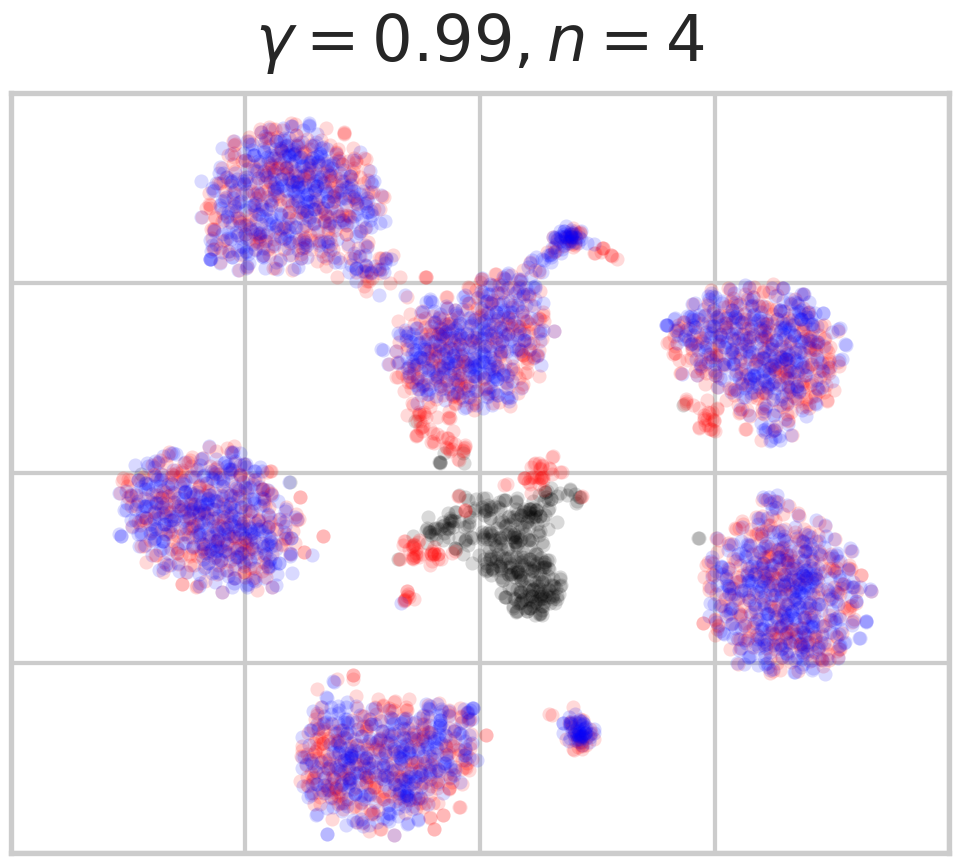} \\
    %\vspace{8pt}
    \includegraphics[width=0.45\linewidth]{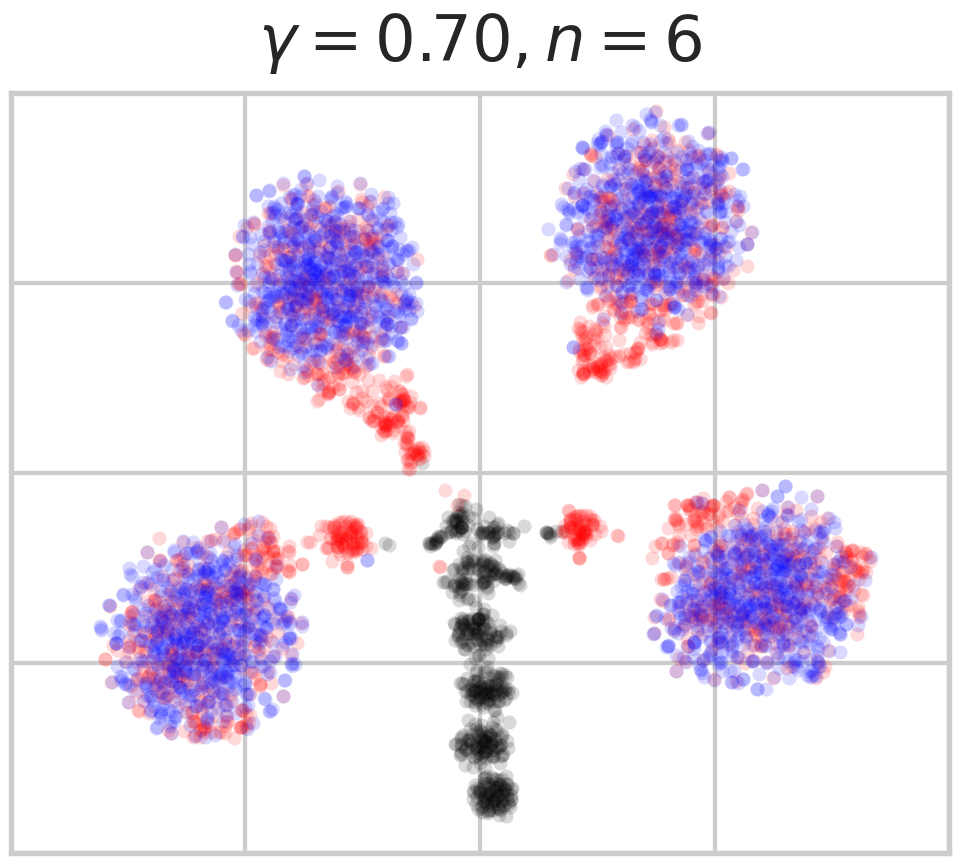}
    \hspace{6pt}
    \includegraphics[width=0.45\linewidth]{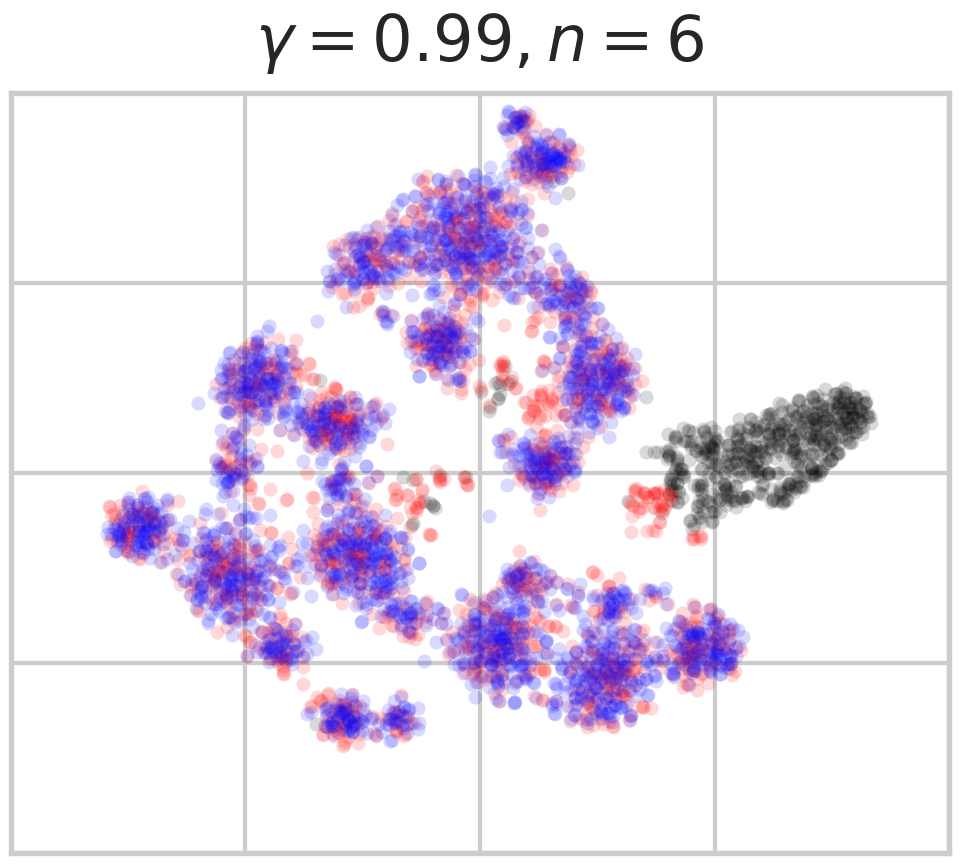}
    \caption{States at different time steps visualised with 2D t-SNE embedding. Black, red and blue points indicate initial, precedent and successor states, respectively.}
    \label{fig:t-sne}
  \end{figure}

  To verify the occurrence of SSC in OLG via RL, particularly in EDRL, we conduct a case study on the widely used SMB level generation task. Several EDRL designers are trained in a finite horizon for one trial, and tested on generating $100$ levels each. The states that appeared during the test are collected and visualised in Fig. \ref{fig:t-sne}. Details of the experimental study are described as follows.

  The implementation of EDRL in \cite{wang2022fun} is used, as it is the latest work of OLG via RL, to our best knowledge. Following \cite{wang2022fun}, the reward function is formulated to assure the playability of levels and maximise the novelty of newly generated level segments with respect to the $n$ latest segments. 
  The values of hyperparameters $\gamma$ (a common parameter in RL) and $n$ in the reward formulation (cf. Eq. (4) in \cite{wang2022fun}) are varied to train different designers, where $\gamma$ determines the degree of emphasis on future rewards and $n$ determines the number of considered historical segments.
  Eight designers with different combinations of $\gamma \in \{0.7, 0.8, 0.9, 0.99\}$ and $n \in \{4, 6\}$ are trained for $1$ million steps with $h=25$, then tested on generating $100$ levels of $2h$ segments. Specifically, an initial state is composed of $n$ randomly sampled latent vectors, i.e., $n$ segments. Therefore, each level has $n + 2h$ segments.

  We collect states by sliding a window of width $n$ on the generated latent vector trajectories.
  The collected states are divided into three categories: \textit{initial}, \textit{precedent} and \textit{successor} by the step they belong to. More precisely, states at $[0,n-1]$ steps are grouped into the \textit{initial} category, states at $[n,h-1]$ steps are grouped into the \textit{precedent} category, while states at the $[h,2h]$ steps are grouped into the \textit{successor} category. 
  We then embed those states with t-SNE \cite{van2008visualizing}, which returns a set of \textit{ad hoc} 2D vectors with similar ratios of Euclidean distances among samples.
  We compute the distances among latent vectors rather than level segments to measure the similarities among the states as they are being directly input into the RL designer, despite there is no guarantee that similar latent vectors correspond to similar level segments.
  Finally, the obtained 2D vectors are visualised with scatter plots. 
  
  Fig. \ref{fig:t-sne} illustrates the results of four representative designers with training hyperparameter settings of $\{\gamma = 0.70, n =4\}$; $\{\gamma = 0.99, n = 4\}$; $\{\gamma = 0.70, n = 6\}$; $\{\gamma = 0.99, n = 6\}$. 
  The blue points are overlapped by red points, therefore, we posit $\S^\pi_{h:2h} \subseteq \S^\pi_{n:h-1} \Rightarrow \S^\pi_{h} \subseteq \S^\pi_{n:h-1}$, and thereby SSC occurs at $[n, h]$ in all the four training settings.
  The red points that appear separated from blue groups are the states in the early stage of the online generation process.
  Another interesting finding is that the scatter plot of the designer trained with $\gamma=0.99$ and $n=6$ (bottom right of Fig. \ref{fig:t-sne}) shows complex patterns compared with the others. It indicates the levels generated by this designer are possibly more diverse. Later evaluation results (cf. Tab. \ref{tab:ablation}) also show that this designer performs better in inter-level diversity.

\section{Quality \textit{v.s.} Diversity through the Lens of SSC\label{sec:exp}}
\begin{figure}[htbp]
    \centering
    \includegraphics[width=0.49\linewidth]{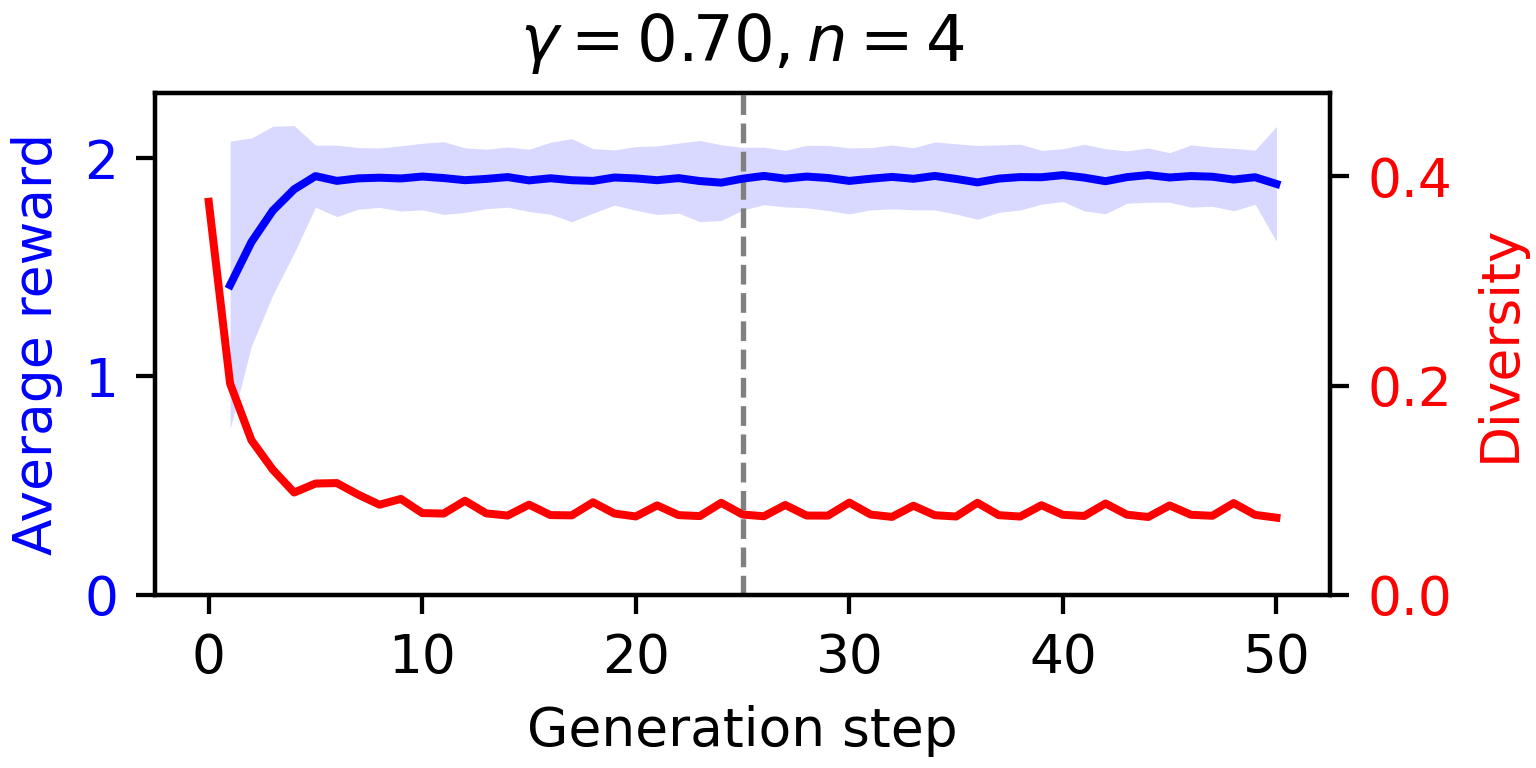}
    \includegraphics[width=0.49\linewidth]{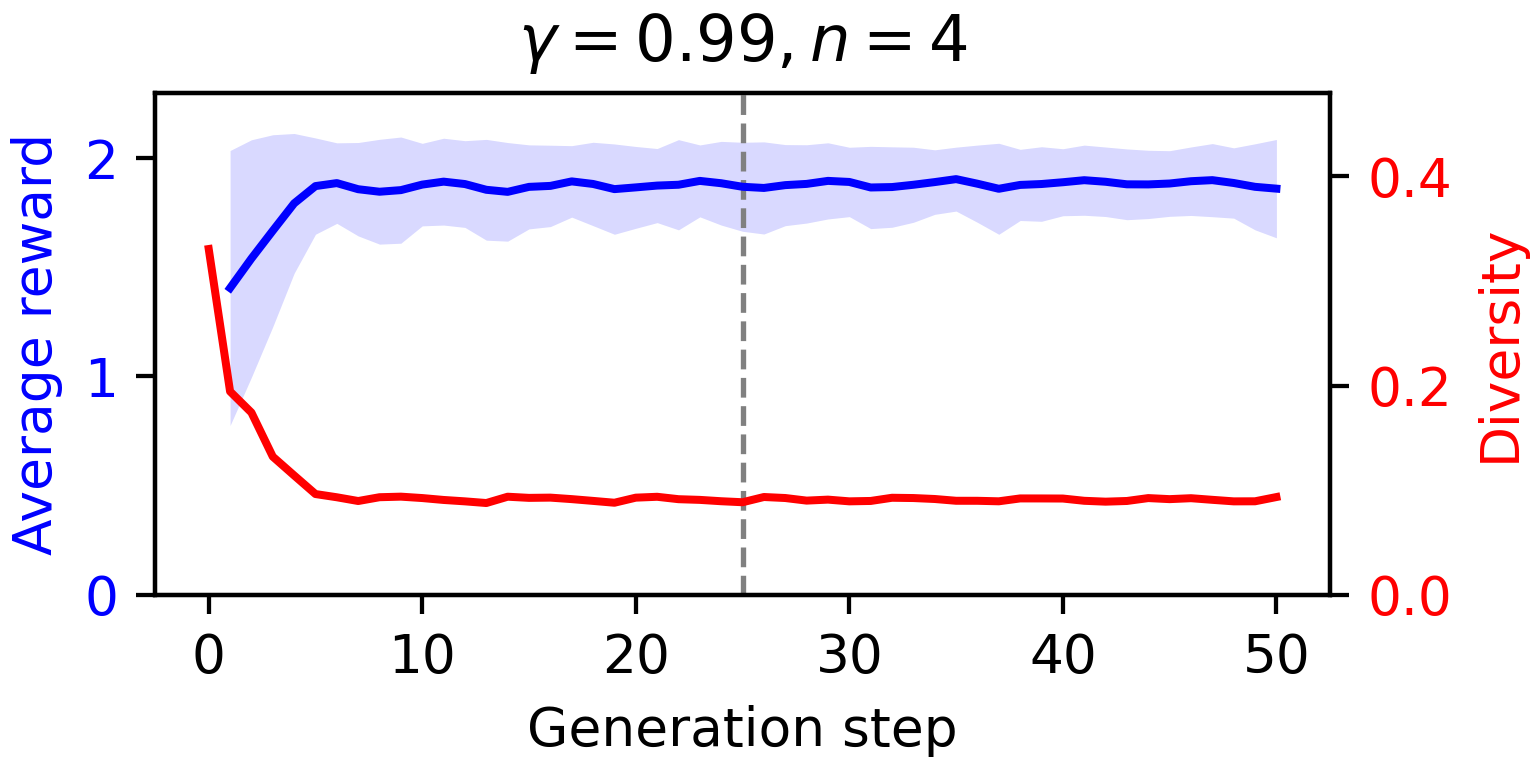}
    \includegraphics[width=0.49\linewidth]{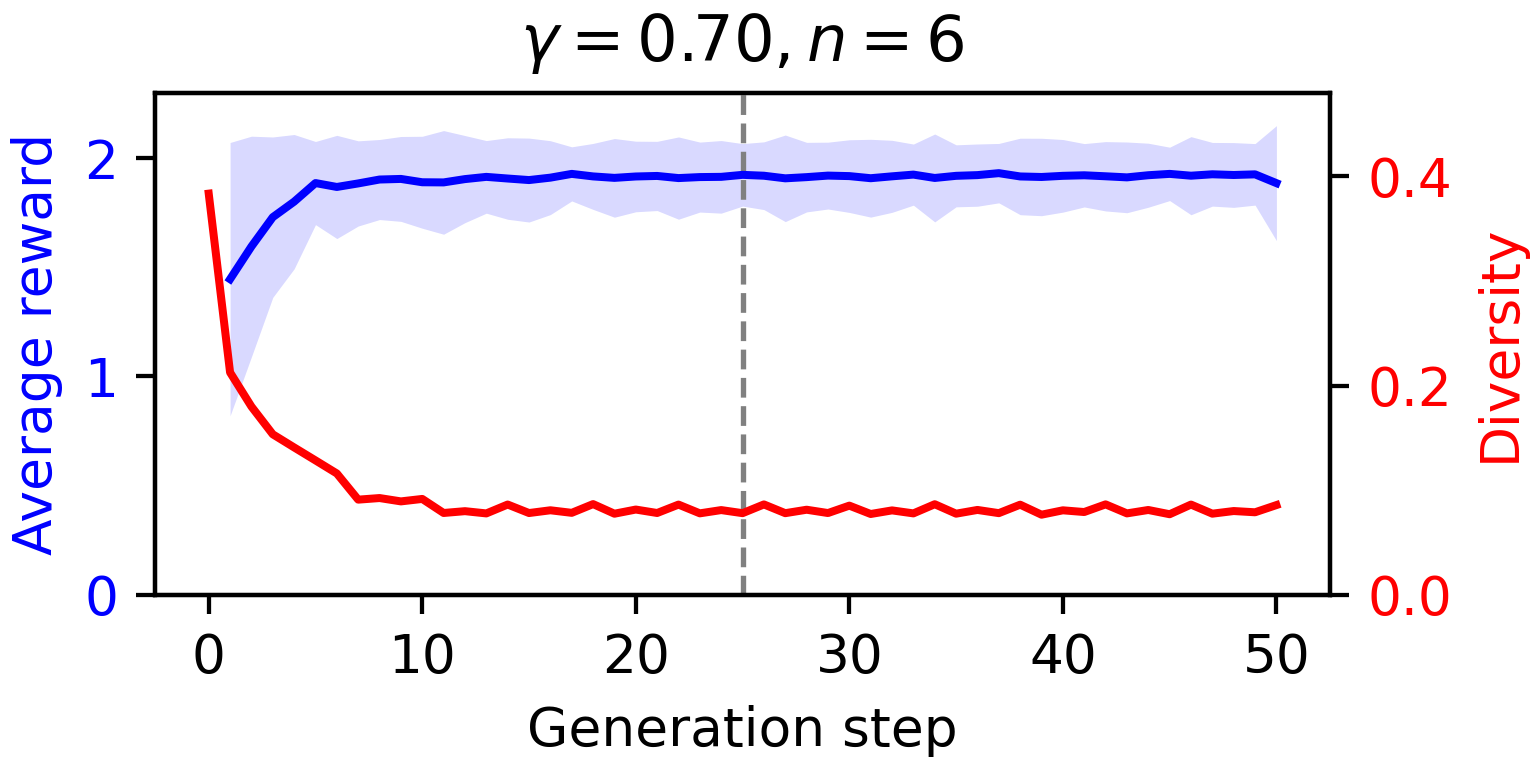}
    \includegraphics[width=0.49\linewidth]{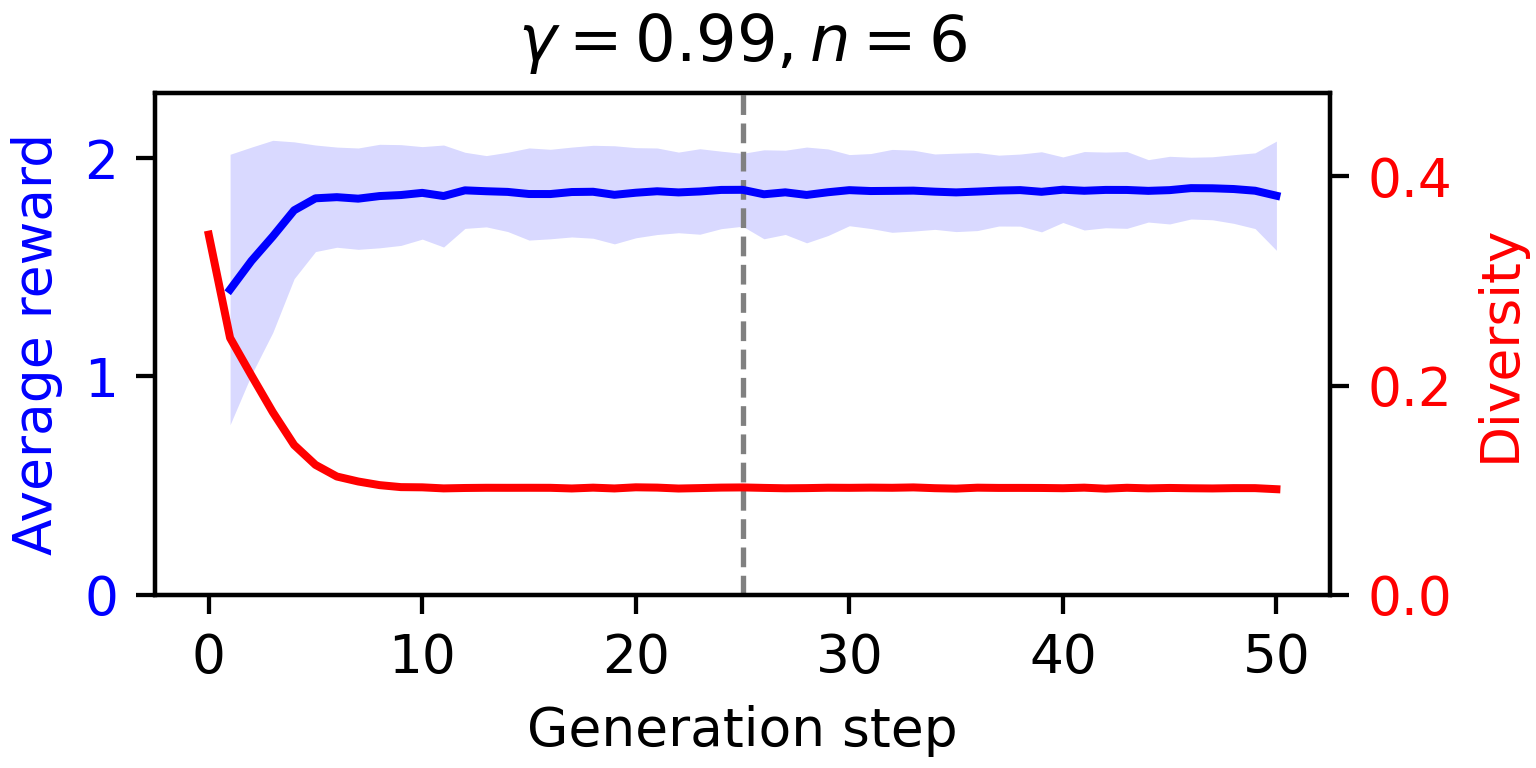}
    \caption{Reward and diversity with respect to time steps. Gray dashed line indicates the number of time steps used for training. Standard deviation is presented using shadow.}
    \label{fig:curves}
  \end{figure}

  As discussed above, SSC is likely to influence the diversity of generated levels, while also assuring the robustness of RL designer for generating long level. These two expectations motivate us to test the quality (reward) and diversity of levels generated by EDRL with respect to the number of generation steps. Taking again the four designers shown in Fig. \ref{fig:t-sne} as examples, we plot the rewards and diversity of segments generated by each designer with respect to the generation step. The diversity is evaluated among all segments generated at the $\ist{i}$ step for each $i \in \{1, 2, \dots, 2h\}$, via the \textit{minimal neighbour distance} proposed by Preuss \etal \cite{preuss2014searching}. Formally, minimal neighbour distance can be expressed as
  \begin{equation}
    \mnd(X|\Re) = \frac{1}{|X|} \sum_{x \in X} \min_{x' \in \Re} D(x, x'), 
    \label{eq:mnd}
  \end{equation} 
  where $X$ is the sample set to be evaluated, $\Re$ is a set of references, and $D(\cdot, \cdot)$ is a distance measurement, which is Euclidean distance in our case. Specifically, the work of \cite{preuss2014searching} determines the references $\Re$ by applying clustering algorithms on the sample set. 
  In this work, we measure the rewards in 1000 trials, and then apply $k$-means \cite{hartigan1979algorithm} with $k=10$ on all the states generated in the test, for each designer. The centroids obtained are recorded as references for each designer. Then minimal neighbour distances at each step $0, 1, \dots, 50$ are calculated for $30$ times to take the average. Each time $100$ levels are generated. The results are shown in Fig. \ref{fig:curves}.

  As shown in Fig. \ref{fig:curves}, there is no considerable deterioration of reward observed from any designer, which supports the outlook that SSC guarantees robust content quality in endless OLG. Meanwhile, the diversity decreases rapidly as the reward increases at the early stage. 
  A possible explanation is that the designers learn to give up gaining high rewards from the initial segments and make the states converge to a small set so that they can ``be lazy". 

  To further analyse the diversity, the mean pairwise distance is tested over $1000$ pairs of levels for all eight designers. 
  The distance is directly computed between levels rather than latent vectors as the players see and play the former rather than the later. 
  The distance between segments is measured by normalised Hamming distance, which is equal to the ratio of tiles that are different in the two levels. 
  To avoid overestimating the distance between levels like \texttt{ABCABC} and \texttt{BCABCA}, the Hamming distance is plugged into dynamic time warping \cite{berndt1994using}, which computes the shortest distance mapping between the sequence entries. This distance measurement is denoted as $\dtwh(\cdot, \cdot)$. As an example, the straight-forward Hamming distance between \texttt{ABCABC} and \texttt{BCABCA} is $6$ while using dynamic time warping returns $2$. 
  As a reference, a naive designer which always takes action by sampling from the noise distribution of GAN directly, is analysed similarly. Referring to the EDRL generator to be measured as $X$ and the generator using the naive designer as $X_{\textrm{naive}}$, the diversity measurement is formulated as follows.
  \begin{equation}\label{eq:div}
    \mathrm{Div}(X) = \frac{\mathbb{E}[\dtwh(x, x') ~|~ x, x' \sim X]}{\mathbb{E}[\dtwh(x, x') ~|~ x, x' \sim X_{\textrm{naive}}]}.
  \end{equation}
  We also test the average rewards at $[1, 10]$,  $[11, 25]$,  $[25, 50]$ step intervals, respectively. Results of diversity and reward are summarised in Tab. \ref{tab:ablation}.

  According to Tab. \ref{tab:ablation}, the diversity is influenced by both $\gamma$ and $n$. The designer with the largest $\gamma$ performs the best for both $n=4$ and $n=6$, while for each $\gamma$ value, the diversity of the designer with $n=6$ is better than the designer with $n=4$. The reason behind this may be summarised as that both the emphasis on future rewards and the long memory of earlier segments promote the complexity of state transition. 
  Though the designer trained with $\gamma=0.99$ and $n=6$ performs the best in terms of diversity, the diversity is only $45\%$ of the one obtained by the naive designer.
  We consider that EDRL will lose the diversity when achieving high reward by solely using the familiar segments. 

    \setlength\tabcolsep{3.5pt}
  \begin{table}[t]
    \centering
    \caption{Diversity ($\mathrm{Div}$ defined in Eq. \eqref{eq:div}) and average rewards (tested for 1000 trials) at $[1,10]$, $[11, 25]$ and $[26,50]$ step intervals, denoted as $\bar R_{1:10}$, $\bar R_{10:25}$, and $\bar R_{26:50}$, respectively.}
    \label{tab:ablation}
    \begin{tabular}{c|cccc|cccc}
      \toprule[1pt]
       & \multicolumn{4}{c|}{$n=4$} & \multicolumn{4}{c}{$n=6$} \\
      $\gamma$ & $0.70$ & $0.80$ & $0.90$ & $0.99$ & $0.70$ & $0.80$ & $0.90$ & $0.99$ \\
      \midrule[0.75pt]
      $\bar R_{1:10}$ & \textbf{1.810} & 1.805 & 1.804 & \textit{1.760} & 1.791 & \textbf{1.802} & 1.797 & \textit{1.728} \\
      $\bar R_{11:25}$ & 1.903 & \textbf{1.928} & 1.886 & \textit{1.874} & 1.911 & \textbf{1.912} & 1.897 & \textit{1.844} \\
      $\bar R_{26:50}$ & 1.909 & \textbf{1.925} & 1.891 & \textit{1.882} & \textbf{1.917} & 1.914 & 1.898 & \textit{1.849} \\
      \midrule[0.5pt]
      $\mathrm{Div}$ & 0.311 & \textit{0.273} & 0.283 & \textbf{0.364} & 0.346 & \textit{0.328} & 0.344 & \textbf{0.445} \\
      \bottomrule[1pt]
  \end{tabular}
    \end{table}

  Tab. \ref{tab:ablation} also shows the robustness of EDRL for generating long quality level as $\bar R_{26:50}$ does not notably differ from $\bar R_{10:25}$. 
  Small $\gamma$ (compared with the typical recommendation of about $\gamma = 0.99$ in vast RL literature like \cite{haarnoja2018soft}) assures better performance on the reward in our tested OLG task. A conflict between quality and diversity is observed from the results, although the reward function guides designers to moderate intra-level divergence. It may be explained that low diversity makes the designer ``practice'' unvarying generation strategy solely so that higher rewards are gained.

\section{Discussion and Outlook \label{sec:discussion}}
\balance
  We have formulated SSC and examined the quality and diversity of EDRL designers. Experiments show that although the current EDRL approach can online generate high-quality levels robustly over a long time, its ability in generating diverse levels is limited because of SSC.
  
  According to the formulation of SSC, it is deduced that the diversity is mostly influenced by the scale of the closure. In the optimal case, the closure is equal to the whole state space so that EDRL designer can achieve the best diversity degree.
  We argue that the issue of limited diversity can hardly be solved by solely tuning or reformulating the reward function within the current framework. Future work should consider making the state transition more complex, so that the closure can be larger and the diversity can be improved. 

  On the other hand, SSC is likely to assure the robustness of generating high-quality levels endlessly. Though SSC has occurred in the current EDRL implementation, we can not explain how EDRL learns to evoke SSC. Future work investigating OLG via RL are suggested to carefully design the state space and state transition to allow the occurrence of SSC. As the formulation and derivation of SSC are generic, it is interesting to verify in other RL tasks (e.g., \textit{BipedalWalker} in Gym \cite{brockman2016openai}) whether SSC appears and enables robust long-term reward or not.

\section{Conclusion}
  This paper formulates a notion of \textit{state space closure} (SSC), and evaluates both the quality and the diversity of levels generated by EDRL. Based on the theoretical analysis of SSC, we discover that the diversity of the EDRL generator can be strongly related to the scale of SSC. Experimental evaluations of diversity also show that the current EDRL approach makes SSC converge to a relatively small set, which limits the diversity of generated levels. Summing up the analysis and evaluations, the issue of inter-level diversity is revealed. We expect this issue can hardly be solved by solely tuning the reward function and hyperparameters of current approaches.

  Though SSC can limit the level diversity, it also has an elegant property (cf. Property \ref{property}) which allows RL designers to learn all the requisite strategies in endless OLG process, from a finite-horizon MDP. The reward (i.e., level quality) will not decrease when the generation process becomes longer than the one specified in training. We suggest future work on endless OLG via RL to consider if SSC occurs or not. 

  \bibliographystyle{IEEEtran}
  \bibliography{refs}

% Generated by IEEEtran.bst, version: 1.14 (2015/08/26)
\begin{thebibliography}{10}
\providecommand{\url}[1]{#1}
\csname url@samestyle\endcsname
\providecommand{\newblock}{\relax}
\providecommand{\bibinfo}[2]{#2}
\providecommand{\BIBentrySTDinterwordspacing}{\spaceskip=0pt\relax}
\providecommand{\BIBentryALTinterwordstretchfactor}{4}
\providecommand{\BIBentryALTinterwordspacing}{\spaceskip=\fontdimen2\font plus
\BIBentryALTinterwordstretchfactor\fontdimen3\font minus
  \fontdimen4\font\relax}
\providecommand{\BIBforeignlanguage}[2]{{%
\expandafter\ifx\csname l@#1\endcsname\relax
\typeout{** WARNING: IEEEtran.bst: No hyphenation pattern has been}%
\typeout{** loaded for the language `#1'. Using the pattern for}%
\typeout{** the default language instead.}%
\else
\language=\csname l@#1\endcsname
\fi
#2}}
\providecommand{\BIBdecl}{\relax}
\BIBdecl

\bibitem{summerville2018procedural}
A.~Summerville, S.~Snodgrass, M.~Guzdial, C.~Holmg{\aa}rd, A.~K. Hoover,
  A.~Isaksen, A.~Nealen, and J.~Togelius, ``Procedural content generation via
  machine learning ({PCGML}),'' \emph{IEEE Transactions on Games}, vol.~10,
  no.~3, pp. 257--270, 2018.

\bibitem{liu2021deep}
J.~Liu, S.~Snodgrass, A.~Khalifa, S.~Risi, G.~N. Yannakakis, and J.~Togelius,
  ``Deep learning for procedural content generation,'' \emph{Neural Computing
  and Applications}, vol.~33, no.~1, pp. 19--37, 2021.

\bibitem{shu2021experience}
T.~Shu, J.~Liu, and G.~N. Yannakakis, ``Experience-driven {PCG} via
  reinforcement learning: A {Super Mario Bros} study,'' in \emph{Conference on
  Games}.\hskip 1em plus 0.5em minus 0.4em\relax IEEE, 2021, pp. 1--9.

\bibitem{goodfellow2014gan}
I.~J. Goodfellow, J.~Pouget-Abadie, M.~Mirza, B.~Xu, D.~Warde-Farley, S.~Ozair,
  A.~Courville, and Y.~Bengio, ``Generative adversarial nets,'' in
  \emph{Advances in Neural Information Processing Systems}.\hskip 1em plus
  0.5em minus 0.4em\relax MIT Press, 2014, pp. 2672--2680.

\bibitem{volz2018evolving}
V.~Volz, J.~Schrum, J.~Liu, S.~M. Lucas, A.~Smith, and S.~Risi, ``Evolving
  {Mario} levels in the latent space of a deep convolutional generative
  adversarial network,'' in \emph{Genetic and Evolutionary Computation
  Conference}.\hskip 1em plus 0.5em minus 0.4em\relax ACM, 2018, pp. 221--228.

\bibitem{sutton2018reinforcement}
R.~S. Sutton and A.~G. Barto, \emph{Reinforcement learning: An
  introduction}.\hskip 1em plus 0.5em minus 0.4em\relax MIT Press, 2018.

\bibitem{wang2022fun}
Z.~Wang, J.~Liu, and G.~N. Yannakakis, ``The fun facets of {Mario}:
  Multifaceted experience-driven {PCG} via reinforcement learning,'' in
  \emph{International Conference on the Foundations of Digital Games}.\hskip
  1em plus 0.5em minus 0.4em\relax ACM, 2022, pp. 1--8.

\bibitem{khalifa2020pcgrl}
A.~Khalifa, P.~Bontrager, S.~Earle, and J.~Togelius, ``{PCGRL}: Procedural
  content generation via reinforcement learning,'' in \emph{AAAI Conference on
  Artificial Intelligence and Interactive Digital Entertainment}, vol.~16,
  no.~1.\hskip 1em plus 0.5em minus 0.4em\relax AAAI, 2020, pp. 95--101.

\bibitem{werneck2020generating}
M.~Werneck and E.~W. Clua, ``Generating procedural dungeons using machine
  learning methods,'' in \emph{Brazilian Symposium on Computer Games and
  Digital Entertainment}.\hskip 1em plus 0.5em minus 0.4em\relax IEEE, 2020,
  pp. 90--96.

\bibitem{susanto2021applying}
E.~K. Susanto and H.~Tjandrasa, ``Applying hindsight experience replay to
  procedural level generation,'' in \emph{East Indonesia Conference on Computer
  and Information Technology}.\hskip 1em plus 0.5em minus 0.4em\relax IEEE,
  2021, pp. 427--432.

\bibitem{zakaria2022procedural}
Y.~Zakaria, M.~Fayek, and M.~Hadhoud, ``Procedural level generation for
  {Sokoban} via deep learning: An experimental study,'' \emph{IEEE Transactions
  on Games}, 2022, doi: \url{10.1109/TG.2022.3175795}.

\bibitem{wang2022online}
Z.~Wang and J.~Liu, ``Online game level generation from music,'' in
  \emph{Conference on Games}.\hskip 1em plus 0.5em minus 0.4em\relax IEEE,
  2022, pp. 119--126.

\bibitem{van2008visualizing}
L.~van~der Maaten and G.~Hinton, ``Visualizing data using t-{SNE},''
  \emph{Journal of Machine Learning Research}, vol.~9, no.~86, pp. 2579--2605,
  2008.

\bibitem{preuss2014searching}
M.~Preuss, A.~Liapis, and J.~Togelius, ``Searching for good and diverse game
  levels,'' in \emph{Conference on Computational Intelligence and Games}.\hskip
  1em plus 0.5em minus 0.4em\relax IEEE, 2014, pp. 1--8.

\bibitem{hartigan1979algorithm}
J.~A. Hartigan and M.~A. Wong, ``A k-means clustering algorithm,''
  \emph{Journal of the Royal Statistical Society}, vol.~28, no.~1, pp.
  100--108, 1979.

\bibitem{berndt1994using}
D.~J. Berndt and J.~Clifford, ``Using dynamic time warping to find patterns in
  time series.'' in \emph{Knowledge Discovery and Data Mining Workshop},
  vol.~10, no.~16.\hskip 1em plus 0.5em minus 0.4em\relax ACM, 1994, pp.
  359--370.

\bibitem{haarnoja2018soft}
T.~Haarnoja, A.~Zhou, P.~Abbeel, and S.~Levine, ``Soft actor-critic: Off-policy
  maximum entropy deep reinforcement learning with a stochastic actor,'' in
  \emph{International Conference on Machine Learning}.\hskip 1em plus 0.5em
  minus 0.4em\relax PMLR, 2018, pp. 1861--1870.

\bibitem{brockman2016openai}
G.~Brockman, V.~Cheung, L.~Pettersson, J.~Schneider, J.~Schulman, J.~Tang, and
  W.~Zaremba, ``{OpenAI} gym,'' \emph{arXiv preprint arXiv:1606.01540}, 2016.

\end{thebibliography}

\end{document}